\documentclass[10pt,onecolumn]{article}

\usepackage[margin=0.75in]{geometry}
\usepackage{amsmath, amssymb}
\usepackage{graphicx}
\usepackage{booktabs}
\usepackage{cite}
\usepackage{hyperref}
\usepackage{caption}
\usepackage{cite}
\usepackage{algorithm}
\usepackage{algpseudocode}
\usepackage{multirow}

\title{A Kinematic Analysis of Palm Degrees of Freedom for Enhancing Thumb Opposability in Robotic Hands}

\author{
    HyoJae Kang$^{1}$, Yeong Jae Park$^{1}$, Joonho Lee$^{1}$, Hyunmok Jung$^{1}$, and Dong Il Park$^{1,*}$ \\
    $^{1}$Advanced Robotics Research Center, Korea Institute of Machinery \& Materials (KIMM) \\
    $^{*}$Corresponding author
}

\begin{document}
\date{}
\maketitle

\thispagestyle{plain}
\footnotetext{This manuscript has been submitted for possible publication.}

\begin{abstract}
This study investigates the kinematic role of palm degrees of freedom (DoF) in enhancing thumb opposability in a five-finger robotic hand. A hand model consisting of a five DoF thumb and four fingers with three to four DoF is analyzed, where palm motion is introduced between adjacent fingers. To quantitatively evaluate thumb–finger interaction, the overlap workspace volume is defined based on voxelized fingertip reachable regions. Seven cases are considered, including configurations with increased total DoF and configurations in which the total DoF is maintained by redistributing DoF from the fingers to the palm. The results show that palm DoF significantly improves opposability, particularly for the ring and little fingers, by repositioning their base locations rather than simply extending their reachable range. However, when the total DoF is constrained, redistributing DoF to the palm leads to trade-offs between overlap workspace expansion and kinematic redundancy. These findings indicate that palm DoF and finger DoF play distinct roles in hand kinematics and should be considered jointly in design. This study provides a quantitative framework for evaluating palm-induced opposability without relying on object or contact models and offers practical design guidelines for incorporating palm motion in robotic hands.
\end{abstract}

\section{Introduction}

The human hand enables a wide range of interactions with the environment through its complex kinematic structure and functional capabilities\cite{b1,b2}. It is regarded as one of the most sophisticated biological systems for performing both power and precision tasks\cite{b3}. Inspired by these characteristics, robotic systems have developed end-effectors that enable interaction with objects in diverse environments\cite{b4}. These systems have also been applied to repetitive and hazardous tasks\cite{b5,b6,b7}.

To achieve such functionalities, various types of robotic end-effectors have been developed, ranging from low DoF two-finger grippers to multi-finger and anthropomorphic robotic hands. Recently, there has been an increasing demand for robotic systems capable of operating in human-centered environments, where collaboration with humans and handling of objects with diverse geometries are required\cite{c15,b8}. In such scenarios, achieving a high level of dexterity becomes a critical requirement.

Dexterity is commonly used to describe the skillfulness of robotic hands. Previous studies\cite{b9} classify dexterity into three main categories: potential dexterity, grasp dexterity, and manipulability dexterity. Among these, potential dexterity refers to the set of reachable configurations without considering object interaction and is typically characterized by kinematic redundancy and thumb opposability\cite{b9}. Kinematic redundancy represents the number of joint configurations that can achieve a given pose, while thumb opposability can be evaluated through overlap regions between the thumb and other fingers\cite{b10,b11} or through standardized tests such as the Kapandji test\cite{b12}. 

The thumb plays a crucial role in hand dexterity because it moves in a direction distinct from the other fingers and can perform multiple functions\cite{c2}. Previous studies have shown that a large portion of the human hand’s prehensile capability originates from the evolution of thumb opposition/reposition (O/R) motion and the ability of the palm to deform or bend\cite{d36,d37,d38}. Accordingly, to achieve thumb opposition and human-like grasping motions in robotic hands, it is important to establish appropriate hand kinematics while ensuring suitable force distribution among the fingers\cite{d20}.

Various studies have attempted to improve robotic hand dexterity by incorporating palm motion into the hand design. The motion of the human palm arises from relative motion between the metacarpal bones, enabled by the mobility of the carpometacarpal (CMC) joints and supported by intermetacarpal articulations, allowing the palm to adapt its shape\cite{c2}. Palm motion is typically realized through F/E motion of the metacarpals and the opposition of the thumb with respect to the other fingers, enabling dexterous grasping behaviors\cite{c1,c2,c3}.

Based on these characteristics, several robotic designs have implemented palm motion at different metacarpal regions, such as between the little–ring and ring–middle fingers, or across multiple regions simultaneously\cite{c4,c5,c6,c7,c8,c9}. It has been shown that introducing palm mobility can expand the reachable workspace of the fingertips and enhance thumb–finger opposability\cite{c4}. When implementing such palm mechanisms, it is important to consider the range of motion of each digit based on human hand characteristics. The mobility of the second and third CMC joints is relatively limited compared to that of the fourth and fifth CMC joints\cite{c13}, and the fifth CMC joint allows a larger range of motion than the fourth\cite{c12}. Despite these anatomical insights and various design attempts, there is still no consistent guideline for designing palm joints in robotic hands, and many designs rely on empirical parameter selection\cite{c4,c10}.

Despite these efforts, existing studies primarily focus on demonstrating the effectiveness of palm motion in specific designs, while a systematic understanding of how palm DoFs should be incorporated under practical constraints remains limited. In particular, the trade-off between increasing palm DoFs and redistributing existing finger DoFs has not been quantitatively analyzed, despite its importance in actuator-constrained robotic systems. This lack of structured design guidelines makes it difficult to determine when and how palm motion should be introduced in robotic hand design.

Therefore, this study aims to provide a kinematic analysis of the effect of palm DoF on thumb opposability, focusing on the additional flexion motion generated in the ring and little fingers. The analysis is conducted based on a five-finger robotic hand model and is divided into two cases: (1) increasing the total number of DoF by introducing palm motion, and (2) maintaining the total number of DoF by reducing the finger DoF while adding palm DoF. This distinction is important because, in practical robotic systems, increasing the number of actuators is often constrained by hardware complexity, weight, and cost. Therefore, analyzing both scenarios provides a more realistic understanding of how palm DoF can be utilized in practical designs. Through this comparison, the study provides insights that can serve as design guidelines for robotic hand kinematics and clarifies the role of palm DoF in enhancing opposability. This analysis is particularly important for the ring and little fingers, as they are located farther from the thumb and rely more on palm-induced motion to achieve effective opposition.

\begin{figure}
    \centering
    \includegraphics[width=0.9\columnwidth]{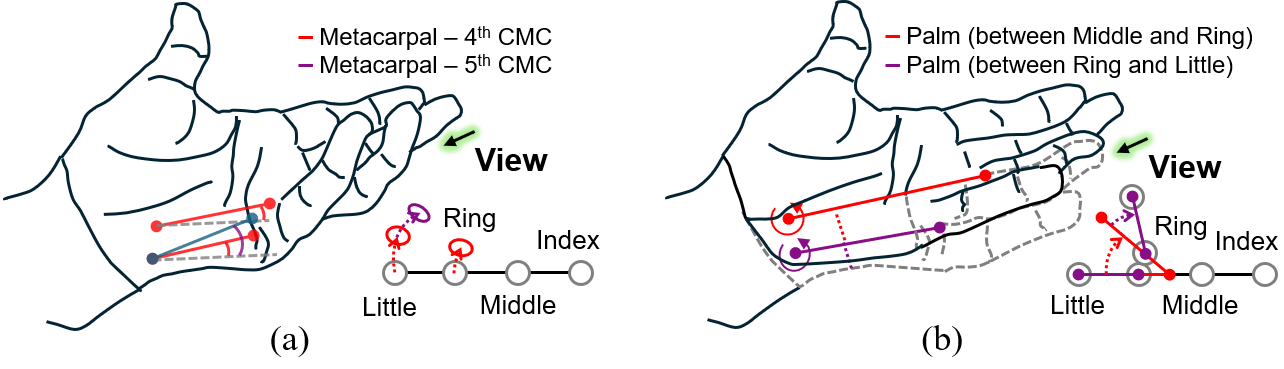}
    \caption{Motion of the palm generated by (a) metacarpal motion, (b) two rotational degrees of freedom between adjacent fingers replacing metacarpal motion}
    \label{fig6}
\end{figure}

In this study, the motion of the metacarpal structure is not modeled to replicate the exact anatomical behavior of the human hand. Instead, it is simplified as a single rotational degree of freedom that effectively pulls the fingers toward the thumb direction, as illustrated in Figure ~\ref{fig6}. Figure ~\ref{fig6}(a) presents the metacarpal motion of the human hand induced by the 4th and 5th CMC joints, whereas Figure ~\ref{fig6}(b) shows the corresponding motion represented as an equivalent rotational degree of freedom between adjacent fingers. This abstraction is introduced to capture the kinematic effect of palm-induced flexion on thumb opposability, rather than to reproduce the detailed biomechanics of the metacarpal joints. By adopting this simplified representation, the influence of palm DoF on finger–thumb interaction can be analyzed in a more tractable and systematic manner.

The scope and contributions of this work are summarized as follows:

\begin{enumerate}
    \item A method for analyzing the effect of palm-induced flexion motion on thumb opposability using overlap workspace volume is proposed.
    \item The influence of increasing palm DoF on the opposability of the ring and little fingers is quantitatively analyzed.
    \item The effect of redistributing DoF—by increasing palm DoF while reducing finger F/E DoF—on thumb opposability is investigated.
\end{enumerate}

The remainder of this paper is organized as follows. Section 2 describes the kinematic structure of the five-finger robotic hand used in this study. Section 3 presents the voxel-based method for computing overlap workspace volume. Section 4 provides the kinematic analysis and results with respect to palm motion. Section 5 discusses the implications of the results, and Section 6 concludes the paper.

\section{Kinematic Structure}

In this study, a five-finger robotic hand composed of four fingers with four DoF each and one thumb with five DoF is considered. The effects of introducing palm DoF between the little and ring fingers, and between the ring and middle fingers, are analyzed. In addition, the effect of reducing the F/E DoF of the ring and little fingers while increasing the palm DoF, resulting in a total of twenty one DoF for the hand, is also investigated.

\begin{figure}[!t]
    \centering
    \includegraphics[width=0.75\columnwidth]{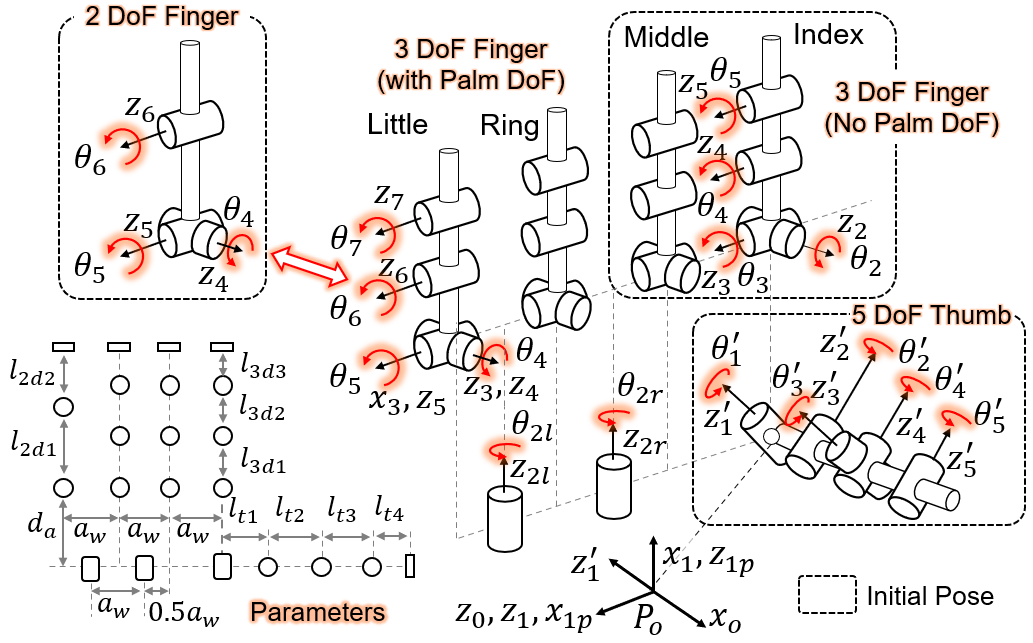}
    \caption{Kinematic structure and parameters of the five-finger robotic hands with five DoF for thumb, three and two DoF for finger, and two DoF for palm flexion motion}
    \label{fig1}
\end{figure}

Figure ~\ref{fig1} shows the kinematic structure of the robotic hand used in the analysis, along with the initial posture and coordinate frames of each finger. With respect to the coordinate frame $P_o x_o y_o z_o$, the first joint coordinate frame of the thumb is denoted by $z_1’$. For the fingers, the first joint coordinate frame is denoted by $z_2$ when there is no palm DoF, by $z_3$ when there is one palm DoF, and by $z_4$ when there are two palm DoFs. When there is no palm DoF, the transformation is defined from $z_o$ to $z_1$. When palm DoFs are introduced, the transformation is instead defined from $z_o$ to $z_{1p}$. The palm DoF between the ring and middle fingers is denoted by $z_{2r}$, and the palm DoF between the ring and little fingers is denoted by $z_{2l}$. In Figure ~\ref{fig1}, the two DoF finger configuration, corresponding to the case where palm DoF is increased and finger DoF is reduced, is also illustrated. The thumb is aligned with the $z_1$ direction of the index finger, and the direction of $z_1’$ defines the rotation axis of its first joint. The dashed lines indicate the initial posture of each finger.

The kinematic parameters between each joint are represented in a skeleton form at the lower left of Figure ~\ref{fig1}. The link lengths of the three DoF finger are denoted as $l_{3d1}$, $l_{3d2}$, and $l_{3d3}$, while those of the two DoF finger are denoted as $l_{2d1}$ and $l_{2d2}$. The link lengths of the thumb are denoted as $l_{t2}$, $l_{t3}$, and $l_{t4}$, and $l_{t1}$ represents the offset between the two CMC joints of the thumb. The distance between adjacent fingers is defined as $a_w$, and the spacing from the little finger to the index finger is uniform. In addition, the palm DoF is located between the two fingers where the motion is generated.

The kinematic parameters are normalized by setting the hand length (HL) to 1. Let the distance from the index finger to the little finger be defined as the hand width (HW). Based on the ratio between the hand width and the overall hand length (HL) reported in \cite{c14}, the average value is determined as follows:

\begin{equation}
HW = 0.54 HL \quad \textrm{and} \quad HW = 3 a_w.
\end{equation}

The thumb lengths, excluding the finger lengths and offset, are divided into equal segments for each phalanx to simplify the analysis. Based on this assumption, the resulting kinematic parameters are summarized in Table ~\ref{tab5}.

\begin{table}[h]
    \centering
    \caption{Kinematic parameters of the five-finger robotic hands}
    \label{tab5}
    \begin{tabular}{c|c|c|c|c}
    $a_w$ & $l_{3d1},l_{3d2},l_{3d3}$ & $l_{t1}$ & $l_{t2}, l_{t3}, l_{t4}$ & $d_a$ \\
    \hline
    0.18 & 0.18 & 0.10 & 0.20 & 0.46 \\
    \end{tabular}
\end{table}

Based on the kinematic structure shown in Figure ~\ref{fig1}, forward kinematics is performed to account for all reachable regions. Let $T_i^j$ denote the homogeneous transformation matrix from frame $i$ to $j$ (where $j > i$). This matrix is obtained through the following sequence of transformations:

\begin{equation}
    T_i^j = T_i^{i+1} T_{i+1}^{i+2}\cdots T_{j-2}^{j-1} T_{j-1}^{j}.
\end{equation}

Each transformation matrix can be derived from the DH parameter tables in Tables ~\ref{tab1}, ~\ref{tab2}, and ~\ref{tab3}. Table ~\ref{tab1} corresponds to the five DoF thumb, and Table ~\ref{tab2} corresponds to the four DoF middle finger. For the other fingers, the representation can be obtained by modifying the offset $a_{i-1}$ in column $i=2$ of the DH table. The DH parameters for the three DoF finger with palm DoF are provided in Table ~\ref{tab3}.

According to the DH parameter convention, the finger joints are modeled as a kinematic chain of interconnected links \cite{c14}. By multiplying the transformation matrices, the transformation from the base frame to the end frame can be obtained. Based on this, the joint positions and fingertip positions for each kinematic structure are expressed with respect to the coordinate frame $P_o x_o y_o z_o$.

\begin{table}[h]
    \centering
    \caption{DH parameters of the five DoF Thumb}
    \label{tab1}
    \begin{tabular}{c|c|c|c|c}
    $i$ & $\alpha_{i-1}$ & $a_{i-1}$ & $d_i$ & $\theta_i$\\
    \hline
    1 & -$\pi$/3 & 0 & 0 & $\theta_1'$ \\
    2 & -$\pi$/2 & $l_{t1}$ & 0 & $\theta_2'$\\
    3 & $\pi$/2 & $l_{t2}$ & 0 & $\theta_3'$\\
    4 & -$\pi$/2 & 0 & 0 & $\theta_4'$\\
    5 & 0 & $l_{t3}$ & 0 & $\theta_5'$\\
    6 & 0 & $l_{t4}$ & 0 & 0\\
    \hline
    \end{tabular}
\end{table}

\begin{table}[h]
    \centering
    \caption{DH parameters of the three DoF middle finger}
    \label{tab2}
    \begin{tabular}{c|c|c|c|c}
    $i$ & $\alpha_{i-1}$ & $a_{i-1}$ & $d_i$ & $\theta_i$\\ \hline
    1 & 0 & 0 & $a_w$ & $\pi$/2 \\
    2 & $\pi$/2 & $d_a$ & 0 & $\theta_2$\\ 
    3 & -$\pi$/2 & 0 & 0 & $\theta_3$\\
    4 & 0 & $l_{3d1}$ & 0 & $\theta_4$\\
    5 & 0 & $l_{3d2}$ & 0 & $\theta_5$\\
    6 & 0 & $l_{3d3}$ & 0 & 0\\
    \hline
    \end{tabular}
\end{table}

\begin{table}[h]
    \centering
    \caption{DH parameters of the three DoF little finger (with two DoF palm)}
    \label{tab3}
    \begin{tabular}{c|c|c|c|c}
    $i$ & $\alpha_{i-1}$ & $a_{i-1}$ & $d_i$ & $\theta_i$\\
    \hline
    1 & -$\pi$/2 & 0 & 0 & -$\pi$/2 \\
    2 & 0 & 1.5$a_w$ & 0 & $\theta_{2r}$\\
    3 & 0 & $a_w$ & 0 & $\theta_{2l}$\\
    4 & 0 & 0.5$a_w$ & $d_a$ & 0\\
    5 & -$\pi$/2 & 0 & 0 & $\theta_4 - \pi$/2\\
    6 & -$\pi$/2 & 0 & 0 & $\theta_5$\\
    7 & 0 & $l_{3d1}$ & 0 & $\theta_6$\\
    8 & 0 & $l_{3d2}$ & 0 & $\theta_7$\\
    9 & 0 & $l_{3d3}$ & 0 & 0\\
    \hline
    \end{tabular}
\end{table}

The initial postures of the thumb and the other fingers are shown in Figure ~\ref{fig1}. In this study, a total of seven cases are defined to analyze the effect of the palm DoF. Table ~\ref{tab4} summarizes the DoF of the palm, ring finger, and little finger for each case. The seven cases are selected to address two representative design questions regarding the use of palm DoF. The first question is whether adding palm DoF can improve thumb opposability when the total number of DoFs is allowed to increase. This is examined using Cases 1 to 4. The second question is whether palm DoF remains beneficial under a fixed total number of DoFs, which is a practical constraint in actuator-limited robotic hands. This is examined using Cases 5 to 7, where finger F/E DoFs are redistributed to the palm while maintaining the same total number of DoFs as Case 1.

Case 1 corresponds to the configuration without palm DoF. Case 2 includes a palm DoF between the little and ring fingers, while Case 3 includes a palm DoF between the ring and middle fingers. Case 4 considers the case with two palm DoFs. Case 5 reduces one F/E DoF of the little finger and introduces a palm DoF between the little and ring fingers. Case 6 reduces one F/E DoF of the ring finger and introduces a palm DoF between the ring and middle fingers. Finally, Case 7 reduces one F/E DoF from both the little and ring fingers and includes two palm DoFs.

\begin{table}[h]
    \centering
    \caption{DoF distribution of the palm, ring finger, and little finger and the location of the palm DoF for each case}
    \label{tab4}
    \begin{tabular}{c|c|c|c}
    Case & Palm DoF & Ring DoF & Little DoF \\
    \hline
    1 & 0 & 3 & 3 \\
    2 & 1(Little) & 3 & 3 \\
    3 & 1(Ring) & 3 & 3 \\
    4 & 2 & 3 & 3 \\
    5 & 1(Little) & 3 & 2 \\
    6 & 1(Ring) & 2 & 3 \\
    7 & 2 & 2 & 2 \\
    \hline
    \end{tabular}
\end{table}

For each of the seven cases, forward kinematics must be derived. The variations in palm DoF and finger DoF can be obtained by modifying Tables ~\ref{tab2} and ~\ref{tab3}. For the case without palm DoF, the configuration can be derived by adjusting the offset in the first row of Table ~\ref{tab2}. For the case with palm DoF, it can be derived by modifying the second to fourth rows of Table ~\ref{tab3}. For the ring finger, only one DoF is applied by excluding $\theta_{2l}$. For the little finger with one palm DoF, the configuration can be obtained by excluding either $\theta_{2l}$ or $\theta_{2r}$ and adjusting the offset accordingly.

The range of motion for each joint angle $\theta$ is defined as follows: for the thumb, it is given in Table ~\ref{tab6}; for the case without palm motion, in Table ~\ref{tab7}; and for the case with palm motion, in Table ~\ref{tab8}. In particular, for the two DoF finger in Table ~\ref{tab8}, $\theta_7$ is not defined. 

\begin{table}[h]
    \centering
    \caption{Thumb Range of motion}
    \label{tab6}
    \begin{tabular}{c|c|c}
    $\theta_1'$ & $\theta_2'$, $\theta_4'$, $\theta_5'$ & $\theta_3'$ \\
    \hline
    0 to $\pi$/2 & -$\pi$/2 to 0 & -$\pi$/6 to $\pi$/6\\ 
    \end{tabular}
\end{table}

\begin{table}[h]
    \centering
    \caption{Finger Range of Motion without Palm Motion}
    \label{tab7}
    \begin{tabular}{c|c|c}
    $\theta_2$ & $\theta_3$ & $\theta_4$, $\theta_5$\\
    \hline
    -$\pi$/12 to $\pi$/12 & -$\pi$/2 to $\pi$/9 & -$\pi$/2 to 0\\
    \end{tabular}
\end{table}

\begin{table}[h]
    \centering
    \caption{Finger Range of Motion with Palm Motion}
    \label{tab8}
    \begin{tabular}{c|c|c}
    $\theta_4$ & $\theta_5$ & $\theta_6$, $\theta_7$\\ 
    \hline
    -$\pi$/12 to $\pi$/12 & -$\pi$/2 to $\pi$/9 & -$\pi$/2 to 0\\ 
    \end{tabular}
\end{table}

The motion associated with the palm DoF is based on the study in \cite{c13}. First, the range of motion for flexion at the fourth CMC joint is defined as follows:

\begin{equation}
    0 \leq \theta_{2r} \leq \frac{\pi}{9}.
\end{equation}

The motion of the 5th CMC joint is the result of the accumulated motion of the 4th CMC joint. When the 4th CMC joint is constrained—i.e., when the palm DoF is driven by the 5th CMC joint—the motion is defined as follows:

\begin{equation}
    0 \leq \theta_{2l} \leq \frac{\pi}{6}.
\end{equation}

When the palm DoF is influenced by both the 4th and 5th CMC joints, the motion of each joint is identical to the case with a single DoF when considered individually. However, when the two motions are combined, the following constraint must be satisfied.

\begin{equation}
    \theta_{2r} + \theta_{2l} \leq \frac{11\pi}{45}.
\end{equation}

The range of motion of the palm is defined based on \cite{c13}. To ensure uniform angular resolution for workspace evaluation, $\theta_{2l}$ is adjusted from 7$\pi$/45 to $\pi$/6. Based on the defined constraints, the motion ranges for the seven cases are determined.

\section{Workspace based evaluation}

The effect of palm motion is analyzed in terms of workspace volume. In particular, this study focuses on thumb opposability to quantitatively evaluate the interaction capability between the thumb and each finger. To this end, the overlap workspace volume between the thumb and each finger is defined as the primary evaluation metric. The workspace volume was approximated by discretizing the Cartesian space into a grid of cubic cells and identifying the cells reached by sampled fingertip positions. The volume was then evaluated based on the total number of such occupied cells~\cite{c41}. Similar voxel-based strategies have been adopted in prior studies to characterize the reachable workspace of robotic manipulators, where the continuous joint space is discretized to generate spatial occupancy representations~\cite{c42}. The overlap workspace refers to the region that both fingers can reach simultaneously, and it can be interpreted as an indicator of the potential capability for grasping and interaction.

In this study, a total of seven cases, as listed in Table ~\ref{tab4}, are analyzed. Case 1 is defined as the baseline configuration without palm DoF. Based on this, Cases 2, 3, and 4 are constructed to analyze the effect of the location and number of palm DoFs. This analysis aims to investigate how the introduction of palm DoF affects thumb opposability from a kinematic perspective.

Next, to exclude the effect of simply increasing the number of DoFs and to compare structural differences under the same total DoF, a condition maintaining the same total DoF as in Case 1 is defined. For this purpose, instead of adding palm DoF, Cases 5, 6, and 7 are defined by reducing the F/E DoF of the fingers at the corresponding locations. This enables a comparative analysis of how the distribution of DoFs affects thumb opposability under the same total number of DoFs.

To compute the workspace volume, the workspace of each finger is defined as the spatial occupancy of fingertip positions generated within the given joint ranges. This represents the reachable space of each finger in a kinematic sense and reflects both the motion range and structural characteristics of the finger. Algorithm ~\ref{alg1} describes the procedure for deriving the workspace volume.

For each finger $f \in {t, i, m, r, l}$, the joint variable vector $\mathbf{q}_f$ is defined as follows.

\begin{equation}
\mathbf{q}_f = [q_{f,1}, q_{f,2}, \cdots, q_{f,m}]^T.
\end{equation}

Here, $t$ denotes the thumb, $i$ the index finger, $m$ the middle finger, $r$ the ring finger, and $l$ the little finger. The dimension of $\mathbf{q}_f$ is determined by the number of joint DoFs of each finger. Each joint is sampled at uniform intervals within its predefined range, forming the sampled joint set $S_f$. This discrete sampling approximates the continuous joint space with a finite set of representative configurations.

For each sample $\mathbf{q}_f \in S_f$, forward kinematics is applied to compute the fingertip position $\mathbf{p}_f \in \mathbb{R}^3$.

\begin{equation}
\mathbf{p}_f = FK_f(\mathbf{q}_f).
\end{equation}

Here, $FK_f$ denotes the forward kinematics mapping based on the kinematic structure of the corresponding finger, which computes the end-effector position for a given joint configuration. The set of all computed fingertip positions can be interpreted as a point set representing the reachable space of the finger.

Since the computed fingertip positions are distributed in a continuous three-dimensional space, voxel-based discretization is applied for efficient representation and comparison. Let the voxel size be $\delta$. Each point is then mapped to a voxel index as follows.

\begin{equation}
\mathbf{k}_f = (\lfloor \frac{x_f}{\delta}\rfloor, \lfloor \frac{y_f}{\delta}\rfloor, \lfloor \frac{z_f}{\delta} \rfloor).
\end{equation}

Here, $\lfloor \cdot \rfloor$ denotes the floor operation, and points with the same voxel index are considered to belong to the same spatial cell. This process converts the point distribution in continuous space into a discrete occupancy representation, where each voxel indicates whether the corresponding region is reachable at least once.

The workspace of each finger is represented as a set of such voxel indices, and duplicate voxels are removed to form a unique voxel index set $V_f$. Accordingly, the workspace volume is computed by multiplying the number of occupied voxels by the volume of a single voxel, as follows.

\begin{equation}
W_i = |V_i|\delta^3.
\end{equation}

Here, a voxel is used as a volumetric discretization unit that partitions the three-dimensional space into uniform cubic cells. The space is represented in a discrete manner based on the occupancy of each cell. This voxel-based representation enables efficient comparison and analysis of the reachable regions of fingers with complex kinematic structures.

\begin{algorithm}[t]
\caption{Workspace Volume Evaluation}
\label{alg1}
\begin{algorithmic}[1]
\Require Joint sample set $\mathcal{S}_f$, forward kinematics function $FK_f(\cdot)$, voxel size $\delta$
\Ensure Workspace voxel set $V_f$, workspace volume $W_f$

\State $V_f \gets \emptyset$

\For{each joint sample $\mathbf{q}_f \in \mathcal{S}_f$}
    \State Compute fingertip position $\mathbf{p}_f \gets FK_f(\mathbf{q}_f)$
    \State Extract coordinates $(x_f, y_f, z_f)$ from $\mathbf{p}_f$
    \State Compute voxel index
    \Statex \hspace{\algorithmicindent} $\mathbf{k}_f \gets \left(\left\lfloor \dfrac{x_f}{\delta} \right\rfloor,\left\lfloor \dfrac{y_f}{\delta} \right\rfloor,\left\lfloor \dfrac{z_f}{\delta} \right\rfloor\right)$
    \State Update voxel set $V_f \gets V_f \cup \{\mathbf{k}_f\}$
\EndFor

\State Compute workspace volume $W_f \gets |V_f| \delta^3$
\State \Return $V_f, W_f$
\end{algorithmic}
\end{algorithm}

Next, to quantitatively evaluate thumb opposability, the overlap workspace volume is defined. The overlap workspace refers to the region that both the thumb and each finger can reach simultaneously, and it is used as a metric representing the interaction capability between two fingers. Algorithm ~\ref{alg2} describes the procedure for computing the overlap workspace volume.

Let $V_t$ denote the workspace voxel set of the thumb, and $V_f$ denote the workspace voxel set of the target finger $f$. The overlap workspace voxel set is then defined as the intersection of the two sets.

\begin{equation}
O_{t,f} = V_t \cap V_f.
\end{equation}

This set represents the spatial region that both the thumb and the corresponding finger can reach, and it reflects the kinematic interaction capability between the two fingers. The overlap workspace volume $O_{t,f}^{\mathrm{vol}}$ is computed by multiplying the number of overlapping voxels by the volume of a single voxel, as follows.

\begin{equation}
O_{t,f}^{\mathrm{vol}} = |O_{t,f}|\delta^3.
\end{equation}

The overlap workspace volume defined in this manner can be used as a kinematic metric to evaluate the interaction capability between fingers without assuming specific objects or contact conditions. In particular, by comparing the overlap workspace between the thumb and each finger, the effect of the presence and distribution of palm DoFs on thumb opposability can be quantitatively analyzed. 

In addition, for each overlap voxel, the number of joint configurations that reach the voxel is counted. A larger value indicates that the same thumb to finger interaction region can be reached through a greater number of alternative joint configurations. 

\begin{algorithm}[t]
\caption{Overlap Workspace Volume Evaluation}
\label{alg:overlap}
\begin{algorithmic}[1]
\Require Thumb workspace voxel set $V_t$, finger workspace voxel sets $V_f$, voxel size $\delta$
\Ensure Overlap voxel sets $O_{t,f}$, overlap volumes $O_{t,f}^{\mathrm{vol}}$

\For{each finger $f \in \{i,m,r,l\}$}
    \State Compute overlap voxel set $O_{t,f} \gets V_t \cap V_f$
    \State Compute overlap volume $O_{t,f}^{\mathrm{vol}} \gets |O_{t,f}| \delta^3$
\EndFor

\State \Return $O_{t,f}, O_{t,f}^{\mathrm{vol}}$
\end{algorithmic}
\end{algorithm}

\section{Analysis and results}

In this section, the predefined settings for the kinematic analysis are described, and the results for different cases are presented. The objective is to investigate the effect of palm DoF and its effect under the condition of maintaining the same total DoF. This analysis is conducted using a total of seven cases.

\subsection{Voxel and resolution}

The voxel size $\Delta$ must be determined for the workspace and overlap workspace analysis. Selecting an appropriate value of $\Delta$ requires a proper resolution, which refers to the discretization level of the joint motion ranges. In this study, both the resolution and $\Delta$ were determined by examining the variation in the results while assigning equal values to all phalanx lengths corresponding to the design variables. The resolution was chosen at the point where the variation decreased below 3\%.

With the total hand length normalized to 1, the thumb phalanx lengths were set to 0.20, 0.20, and 0.20 for the metacarpal, proximal, and distal segments, respectively. For the other fingers, the phalanx lengths were set to 0.18, 0.18, and 0.18 for the proximal, middle, and distal segments, respectively. The workspace volume was evaluated while progressively refining the joint angle resolution starting from $\pi$/18.

In this process, the voxel size was considered at two values, 0.05 and 0.025. Notable changes in the results were observed at resolutions of $\pi$/60 and $\pi$/90 for the thumb and fingers under voxel sizes of 0.05 and 0.025, respectively.

\begin{table}[h] 
\centering
\caption{Comparison of thumb, index finger, and little finger workspace volumes at different resolution - Case 7}
\label{Tab55}
\begin{tabular}{c|cc|cc|cc}

Voxel & \multicolumn{2}{c|}{Thumb} & \multicolumn{2}{c|}{Index Finger} & \multicolumn{2}{c}{Little Finger}\\
($\Delta$)& $\pi$/60 & $\pi$/90 & $\pi$/60 & $\pi$/90 & $\pi$/60 & $\pi$/90 \\
\hline
0.05  & 0.2630 & 0.2639 & 0.0706 & 0.0710  & 0.1215 & 0.1250 \\
0.025 & 0.2137 & 0.2148 & 0.0506 & 0.0508  & 0.0943 & 0.0971  \\
\end{tabular}
\end{table}

Table ~\ref{Tab55} presents the workspace volume with respect to voxel size and resolution. The results show that, for $\Delta=0.05$, the variation in volume remains within 3\% for five DoF thumb, four DoF index finger, and three DoF little finger with two DoF Palm as the resolution changes. As the voxel size decreases, the computed occupied volume also decreases, while larger voxel sizes result in larger volumes even with fewer occupied points. Furthermore, increasing the resolution leads to a larger computed volume due to denser sampling; however, this increase becomes negligible beyond a certain threshold. Accordingly, the voxel size is set to $\Delta = 0.05$, with a resolution of $\pi$/60 applied to both the thumb and the other fingers.

\subsection{Analysis on adding palm DoF}

\begin{figure}[!t]
    \centering
    \includegraphics[width=0.65\columnwidth]{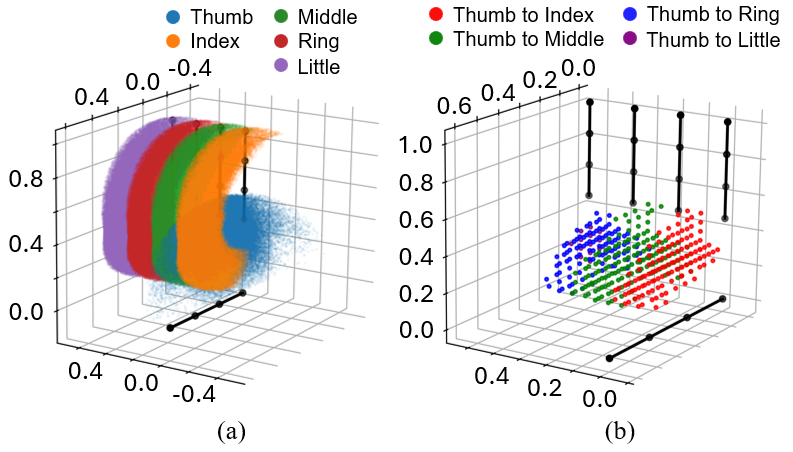}
    \caption{Visualization of the kinematic structure in Case 1 - (a) reachable region, (b) overlap workspace voxel points}
    \label{fig2}
\end{figure}

Before analyzing the effect of the palm DoF, the results for Case 1, where no palm DoF is present, are first examined, as shown in Figure ~\ref{fig2}. Figure \ref{fig2}(a) illustrates the reachable workspace of the proposed hand for each finger. Based on this, the overlap regions between the thumb and the other four fingers are shown in Figure \ref{fig2}(b).

\begin{figure}[!t]
    \centering
    \includegraphics[width=0.6\columnwidth]{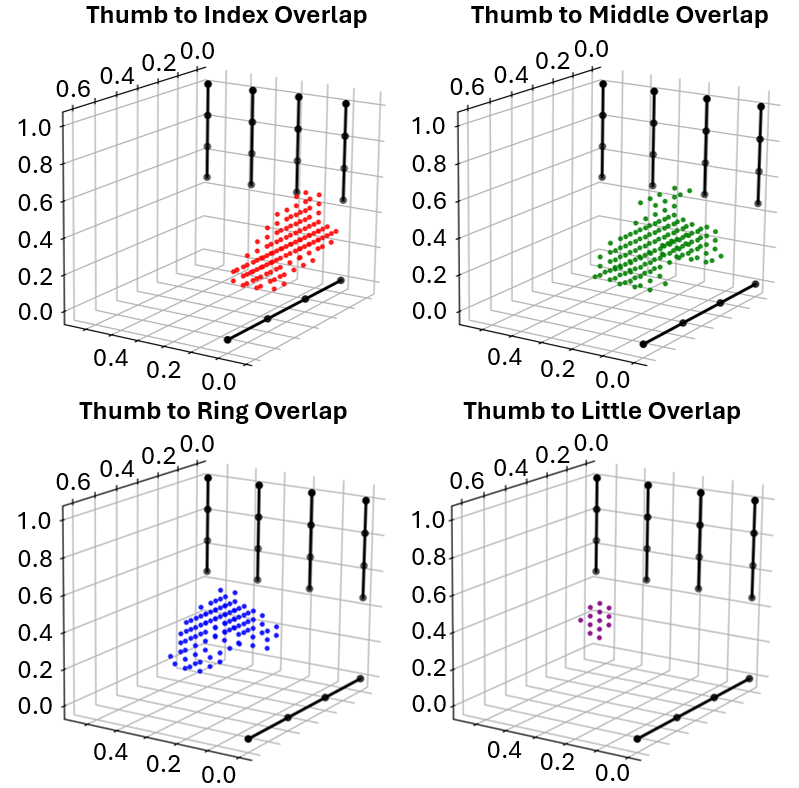}
    \caption{Overlap voxel points detected between the thumb and each finger in Case 1}
    \label{fig3}
\end{figure}

The overlap regions between the thumb and the other four fingers are further distinguished for each finger, as shown in Figure ~\ref{fig3}. Each point represents a single voxel. It can be observed that the overlap workspace decreases in the order of the middle finger, index finger, and ring finger, and the little finger exhibits a relatively smaller overlap compared to the other fingers.

The results corresponding to Figure ~\ref{fig3} are quantified using the detected voxels, as summarized in Table ~\ref{tabA1}. Table ~\ref{tabA1} presents the overlap workspace volumes between the thumb and the other four fingers in Case 1. All four fingers in Case 1 share the same structure, with a maximum workspace volume of 0.069875, while the thumb has a workspace volume of 0.263.

In subsequent cases, the index and middle fingers remain structurally unchanged; therefore, no variation is expected for these fingers. The primary focus is on the ring and little fingers. In Case 1, the ring and little fingers exhibit smaller overlap workspaces compared to the other fingers. Specifically, the ring finger accounts for 18.43\% of its reachable workspace, while the little finger accounts for 2.33\%.

\begin{table}[h]
    \centering
    \caption{Overlap workspace volume based on the number of voxel points - Case 1}
    \label{tabA1}
    \begin{tabular}{c|c|c|c}
    Index & Middle & Ring & Little  \\
    \hline
    0.015625 & 0.019375 & 0.012875 & 0.001625 
    \end{tabular}
\end{table}

\begin{figure}[!t]
    \centering
    \includegraphics[width=0.85\columnwidth]{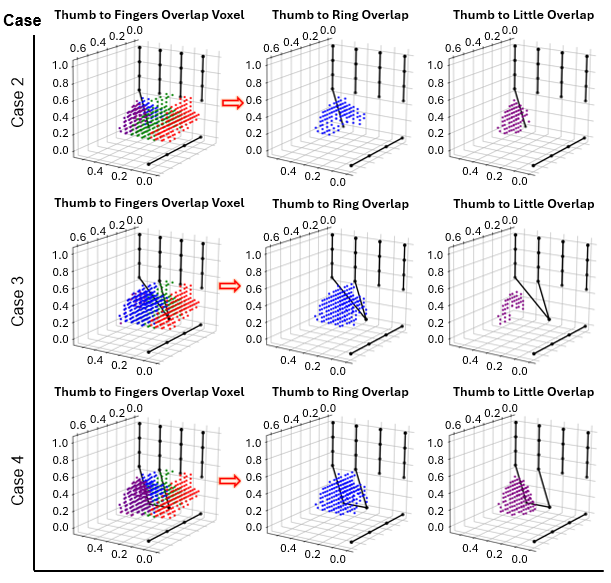}
    \caption{Overlap voxel detection results and overlap voxel points of fingers influenced by palm DoF, Case 2 - palm DoF between the ring and little fingers, Case 3 - palm DoF between the ring and middle fingers, Case 4 - two palm DoFs}
    \label{fig4}
\end{figure}

Next, based on the twenty one DoF hand in Case 1, the results are compared while introducing palm DoF. Figure ~\ref{fig4} shows the overall overlap workspace for Cases 2, 3, and 4, along with the overlap workspaces of thumb-to-ring and thumb-to-little.

In Figure ~\ref{fig4}, Case 2 shows the configuration where one palm DoF is added between the little finger and the ring finger. This results in a twenty two DoF hand. Since the ring finger maintains the same structure as in Case 1, it exhibits the same overlap workspace volume. In contrast, for the little finger, the additional flexion motion toward the thumb direction increases the number of detected voxel points.

In Figure ~\ref{fig4}(b), Case 3 shows the case where one palm DoF is added between the ring finger and the middle finger, resulting in a twenty two DoF hand. Compared to Case 2, although only one additional DoF is introduced, both the ring and little fingers undergo flexion toward the thumb direction. As a result, the number of detected voxel points increases for both the ring and little fingers compared to Case 1.

In Figure ~\ref{fig4}(c), Case 4 shows the configuration where one palm DoF is added between the ring and middle fingers, and another between the ring and little fingers, resulting in a twenty three DoF hand. From the perspective of palm motion, this configuration can be interpreted as a two-link planar structure. Consequently, the palm flexion of the little finger becomes more pronounced than in Cases 2 and 3, allowing it to move closer to the thumb. As a result, the number of detected voxel points for the ring finger is the same as in Case 3, while for the little finger it is greater than in both Cases 2 and 3.

\begin{table}[h]
    \centering
    \caption{Reachable workspace volume based on the number of voxel points for three cases with increasing total DoFs of the hand}
    \label{tabA2}
    \begin{tabular}{c|c|c}
    Case & Ring finger & Little finger \\
    \hline
    2 & 0.069875 & 0.131500 \\
    3 & 0.109750 & 0.121375\\
    4 & 0.109750 & 0.174250 
    \end{tabular}
\end{table}

Table ~\ref{tabA2} presents the reachable workspace volumes of the ring and little fingers in Cases 2, 3, and 4. Based on these results, the corresponding overlap workspace volumes for each case are summarized in Table ~\ref{tabA3}.

\begin{table}[b]
    \centering
    \caption{Overlap workspace volume based on the number of voxel points for three cases with increasing total DoFs of the hand}
    \label{tabA3}
    \begin{tabular}{c|c|c}
    Case & Ring finger & Little finger \\
    \hline
    2 & 0.012875 & 0.010750\\
    3 & 0.025875 & 0.005125\\
    4 & 0.025875 & 0.019000 
    \end{tabular}
\end{table}

In Cases 3 and 4, where palm DoF affecting the ring finger is introduced, the overlap workspace volume increases by 100.97\% in absolute volume compared to Case 1 without palm DoF. The ratio relative to the total reachable workspace also increases from 18.43\% in Case 1 to 37.03\%.

For the little finger, Cases 2, 3, and 4, where palm DoF affecting the little finger is introduced, show different results. In all three cases, the absolute overlap workspace volume increases compared to Case 1, with increases of 561.54\%, 215.38\%, and 1069.23\%, respectively. The ratio relative to the total reachable workspace changes from 2.33\% in Case 1 to 8.17\% in Case 2, 4.22\% in Case 3, and 10.90\% in Case 4.

\subsection{Analysis on adding palm DoF with constrained total DoF}

In many studies on robotic hands, increasing the number of actuated DoFs can improve performance in a wider range of tasks. However, accommodating additional components inside and outside the hand to achieve this is complex and challenging. Therefore, the effect of palm DoF is analyzed while maintaining the same total number of DoFs.

The conditions are briefly summarized as follows. When a palm DoF is introduced between the middle and ring fingers, the F/E DoF of the ring finger is reduced from three DoF to two DoF. When a palm DoF is introduced between the ring and little fingers, the F/E DoF of the little finger is reduced. When two palm DoFs are applied, the F/E DoFs of both the ring and little fingers are reduced.

\begin{figure}[!t]
    \centering
    \includegraphics[width=0.85\columnwidth]{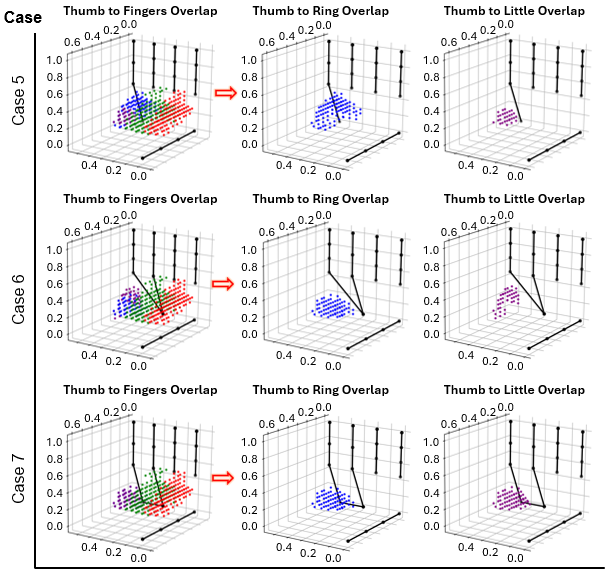}
    \caption{Comparison of reachable workspace, total overlap workspace, and the overlap workspaces of the thumb to ring and little fingers for three cases with varying palm DoF while maintaining the total number of DoFs}
    \label{fig5}
\end{figure}

Reducing the F/E DoF of the fingers affects the reachable workspace volume. Figure ~\ref{fig5} shows the reachable workspace, overlap workspace, thumb-to-ring overlap workspace, and thumb-to-little overlap workspace for Cases 5, 6, and 7.

The reachable workspace volumes of the ring and little fingers in Cases 5, 6, and 7 are summarized in Table ~\ref{tabA4}. In Case 5, the ring finger has the same DoF, reachable workspace volume, and thumb-to-ring overlap as in Case 1. For the little finger, the F/E DoF is reduced from three to two, and one palm DoF is introduced, resulting in a change in the reachable workspace. In Case 6, the little finger has the same DoF, reachable workspace volume, and thumb-to-little overlap as in Case 3. For the ring finger, the reduction in F/E DoF leads to a change in the reachable workspace. In Case 7, the reachable workspace changes compared to Case 4 due to the reduction of F/E DoFs in both the ring and little fingers.

\begin{table}[h]
    \centering
    \caption{Overlap workspace volume based on the number of voxel points for three cases with maintaining total DoFs of the hand}
    \label{tabA4}
    \begin{tabular}{c|c|c}
    Case & Ring finger & Little finger \\
    \hline
    5 & 0.069875 & 0.089588 \\
    6 & 0.071375 & 0.121375 \\
    7 & 0.071375 & 0.121500 \\
    \end{tabular}
\end{table}

Comparing the absolute magnitude of the modified reachable workspace in Cases 5, 6, and 7 with Case 1 is not directly suitable for evaluating the generated motions. Therefore, it is more appropriate to compare Cases 5, 6, and 7 with Cases 2, 3, and 4, respectively. In Case 5, the workspace volume decreases compared to Case 2, and the ratio of overlap workspace volume is 5.37\% for the ring finger and 4.19\% for the little finger. Similarly, in Cases 6 and 7, the workspace volumes decrease compared to Cases 3 and 4. The overlap workspace volume ratio for the ring finger in both Cases 6 and 7 is 12.61\%. For the little finger, the overlap workspace ratio is 4.22\% in Case 6 and 7.00\% in Case 7.

Cases 1 and 5, 6, and 7 have the same total number of DoFs. However, in terms of absolute overlap workspace volume, the ring finger shows a decrease of 30.10\% in both Cases 6 and 7 compared to Case 1. In contrast, the little finger shows an increase of 130.77\% in Case 5, 215.38\% in Case 6, and 423.08\% in Case 7.

\subsection{Voxel-wise number of reachable configurations}

\begin{figure}
    \centering
    \includegraphics[width=0.85\columnwidth]{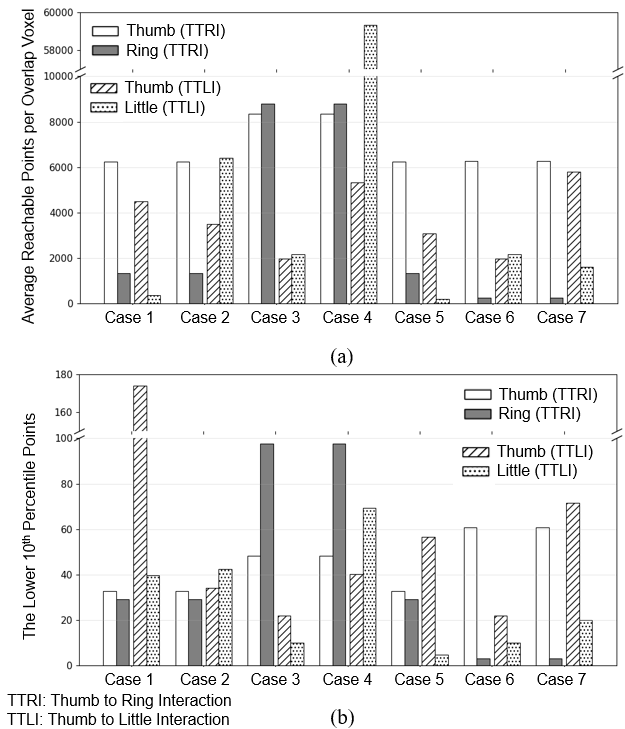}
    \caption{The number of reachable configurations per detected overlap voxel (a) the mean number, (b) the lower 10th percentile for each cases}
    \label{fig7}
\end{figure}

Figure~\ref{fig7} illustrates the number of reachable configurations per detected overlap voxel. Figure~\ref{fig7}(a) shows the mean number of configurations, while Figure~\ref{fig7}(b) presents the lower 10th percentile (p10), representing the lower-bound voxel-wise number of reachable configurations (VWRC). Based on Case 1, which does not include palm degrees of freedom (DoF), the effects of increasing palm DoF can be analyzed in terms of both overlap workspace and VWRC. 

The results of Case 2 include the overlap workspace of Case 1, with additional regions being generated. In Case 2, the overlap workspace of the little finger expands, and both the mean and p10 values increase. This indicates that not only the reachable region is enlarged, but also the VWRC is improved. In contrast, the thumb exhibits a different trend. Although the overlap region expands, both the mean and p10 values decrease, suggesting that the newly added regions are composed of relatively sparsely distributed configurations. Therefore, Case 1 can be interpreted as having a more concentrated distribution of thumb configurations, whereas Case 2 redistributes them over a wider region that includes the original workspace.

Similarly, Case 3 also includes the overlap workspace of Case 1. In Case 3, both the thumb and the ring finger show a significant increase in voxel count and VWRC. The simultaneous increase in the mean and p10 values indicates that the improvement is not limited to specific regions but is uniformly distributed across the overlap workspace. The little finger exhibits a tendency similar to that in Case 2, maintaining the expanded workspace while also increasing VWRC.

Case 4 includes the overlap workspaces of Cases 1 and 3. It shows a similar trend to Case 3 for the ring finger, while the little finger achieves the highest voxel count and VWRC among Cases 1–4. Notably, the p10 value is also maximized, indicating that the improvement is distributed throughout the entire workspace rather than being confined to specific regions. For the thumb, although the configuration concentration is lower than in Case 1, both the mean and p10 values are higher than those in Cases 2 and 3, suggesting a more balanced configuration distribution.

In Case 5, the overlap workspace expands; however, both the mean and p10 values decrease compared to Case 1. This indicates that the newly added regions are composed of more sparsely distributed configurations, leading to reduced effective redundancy. In addition, the reduction of one degree of freedom (DoF) in the little finger compared to Case 2 appears to have a significant impact on this behavior.

In Case 6, the voxel count of the ring finger decreases, while that of the little finger increases. For the thumb–ring pair, the thumb maintains a configuration distribution similar to that of Case 1, whereas the ring finger shows a significant decrease in VWRC. In terms of p10, the little finger decreases, while the thumb becomes more concentrated. For the thumb–little pair, the thumb configurations become more dispersed, whereas the little finger exhibits higher average multiplicity but lower p10, indicating uneven improvement. Compared to Case 3, this behavior can be attributed to the reduction of one DoF in the ring finger.

In Case 7, the ring finger exhibits a trend similar to that of Case 6, while both the thumb and little finger show expanded overlap regions. Compared to Case 6, the thumb shows improvements in both the mean and p10 values, indicating enhanced robustness and a more balanced configuration distribution. In contrast, the little finger shows a decrease in the mean but an increase in p10, suggesting a trade-off between overall multiplicity and lower-bound performance. Compared to Case 4, this behavior can be explained by the reduction of one DoF in both the ring and little fingers.

\subsection{Results}

In this study, the effect of introducing palm DoF on the reachable workspace of the fingers and the overlap workspace with the thumb was analyzed through a total of seven cases. In particular, the effects were compared by distinguishing between cases where the total DoF is increased and cases where it is maintained.

The introduction of palm DoF increases the flexion motion of the ring and little fingers, which are relatively farther from the thumb, resulting in a significant increase in the overlap workspace with the thumb. This can be interpreted as a result of changes in the base orientation of the fingers induced by the palm structure, enabling access to regions that were previously difficult to reach. Consequently, it is confirmed that palm DoF plays an important role in improving the opposability of the hand. In particular, when two palm DoFs are introduced, the flexion motion of the little finger leads to a more pronounced improvement in opposability compared to the case with a single palm DoF.

When palm DoF is introduced while simultaneously reducing the F/E DoF of the fingers, the structure may be advantageous in terms of maintaining the total number of DoFs; however, the results are limited from the perspective of opposability. In the case of the ring finger, although additional motion is introduced by the palm DoF, the reduction in the original F/E motion leads to a decrease in the overlap workspace with the thumb. This indicates that the articulation capability of the finger itself remains an important factor in achieving opposability. In contrast, for the little finger, the effect of palm DoF is more pronounced. In some cases, even with a reduction in F/E DoF, the overlap workspace increases. This is attributed to the relatively larger initial distance between the little finger and the thumb, where changes in the base position induced by palm motion provide greater geometric advantages.

The VWRC analysis provides complementary insight to overlap volume. In Cases 2–4, the introduction of palm DoF increases the overlap voxel count of the ring and little fingers, with Cases 3 and 4 showing improvements in both the mean and p10 values for the thumb–ring pair, while Case 4 yields the most significant improvement for the thumb–little pair. In contrast, in Cases 5–7, where the total DoF is maintained by reducing finger F/E DoF, the expansion of overlap workspace does not consistently lead to improved VWRC, and decreases in mean and p10 values are observed in several cases. It should be noted that this metric does not directly represent task-level manipulation performance, grasp stability, or force capability. Instead, it is used as a supplementary kinematic indicator that describes the distribution and redundancy of reachable configurations within the overlap workspace. Therefore, it is interpreted as a potential dexterity-related measure rather than a direct predictor of actual manipulation success.


\section{Discussions}

The results demonstrate that increasing the palm DoF leads to a consistent improvement in opposability, as evidenced by the increase in overlap workspace between the thumb and the opposing fingers. Unlike finger DoFs, which primarily influence motion within individual fingers, palm DoFs modify the spatial relationships among fingers by repositioning their base locations. This enables fingers that are initially distant from the thumb to move closer, thereby enhancing their ability to interact with the thumb.

In addition to this improvement in overlap workspace, a trade-off between workspace expansion and voxel-wise number of reachable configurations is observed. Increasing palm DoF tends to enlarge the overlap workspace; however, the additional regions do not always maintain high configuration density. The voxel-wise analysis shows that some kinematic structures improve global overlap coverage, while others enhance the local concentration and robustness of reachable configurations. Therefore, overlap volume alone is insufficient to fully explain the structural effect of palm DoF, and the distribution of reachable configurations within the overlap region must also be considered.

From a geometric perspective, thumb length also plays an important role in determining opposability. Increasing the thumb phalanx length can improve overlap with the ring and little fingers, which are located relatively farther from the thumb. However, excessive thumb length may introduce motion constraints due to collisions or interference with the palm structure or other fingers. Therefore, simply extending the thumb does not necessarily improve opposability when the overall hand configuration remains unchanged.

Similarly, the spacing between fingers significantly affects thumb–finger interaction. As the inter-finger distance increases, the thumb must reach farther to interact with the ring and little fingers. This effect becomes more pronounced in larger hands, where both finger lengths and spacing increase proportionally. As a result, the overlap workspace between the thumb and these fingers is reduced, indicating a limitation of scaling-based approaches for improving opposability.

These geometric and kinematic characteristics potentially influence manipulation capability. While three-finger configurations are generally sufficient for stable grasping, more complex manipulation tasks often require multiple simultaneous contact points. In such cases, fingers that are initially distant from the thumb may have limited ability to participate in manipulation. Introducing palm DoFs enables these fingers to be repositioned toward the object, thereby increasing their participation and enabling more effective multi-finger coordination.

Based on these observations, the following design guidelines can be derived. 
\begin{enumerate}
    \item When opposability is limited by the geometric separation between the thumb and distal fingers, introducing palm DoFs is more effective than simply increasing finger length, as it directly repositions the finger bases rather than extending the reach alone. 
    \item The effectiveness of palm DoFs becomes more pronounced as the inter-finger spacing increases or as the overall hand scale grows, indicating that scaling-based approaches alone may be insufficient for maintaining thumb–finger interaction. 
    \item In actuator-constrained designs, redistributing DoFs from individual fingers to the palm can be an effective strategy, particularly for improving the participation of the ring and little fingers in multi-finger interactions. 
\end{enumerate}
    
\section{Conclusions}

This study presented a kinematic analysis of palm DoF in a five-finger robotic hand using workspace volume and overlap workspace volume. To examine the effect of palm motion on thumb opposability, seven cases were defined by varying the number and location of palm DoFs and by additionally considering configurations in which the total DoF of the hand was maintained. The analysis showed that the introduction of palm DoF increases the overlap workspace between the thumb and the fingers affected by palm motion. The main contribution of this study is the quantitative evaluation of palm-induced opposability using voxel-based overlap workspace analysis without assuming object shape or contact conditions. The presented framework makes it possible to compare different palm structures under different kinematic structures and provides a basis for evaluating how palm motion should be incorporated in robotic hand design. 

The limitations of this study are as follows. First, the analysis of palm motion was primarily limited to directions parallel to the fingers. Second, regions that are physically inaccessible due to the thickness and width of the hand were not considered. Third, the analysis considers fingertip position only and does not include fingertip orientation, which may affect contact feasibility and manipulation quality. These factors cannot be neglected in practical robotic hand design. Therefore, in future work, the palm DoF will not be restricted to directions parallel to the fingers, and oblique configurations will also be considered. This will enable the investigation of not only metacarpal-like motion observed in the human hand, but also potential advantages unique to robotic hand designs. In addition, the proposed DoF configurations will be implemented in a physical robotic hand and validated experimentally. Through this, the arrangement of an appropriate number of DoFs will be investigated by considering the usability of fingers located farther from the thumb. In addition, future work will extend the position-based analysis by incorporating fingertip orientation, enabling the evaluation of grasp-related metrics and more realistic manipulation capabilities. Ultimately, the goal is to improve the dexterity of robotic hands by applying suitable palm motion, thereby achieving enhanced manipulation performance.

\vspace{5mm} 

\bibliography{bibtex}
\bibliographystyle{ieeetr}

\end{document}